# WEB SPAM CLASSIFICATION USING SUPERVISED ARTIFICIAL NEURAL NETWORK ALGORITHMS

Ashish Chandra, Mohammad Suaib, and Dr. Rizwan Beg

Department of Computer Science & Engineering, Integral University, Lucknow, India

## ABSTRACT

*Due to the rapid growth in technology employed by the spammers, there is a need of classifiers that are more efficient, generic and highly adaptive. Neural Network based technologies have high ability of adaption as well as generalization. As per our knowledge, very little work has been done in this field using neural network. We present this paper to fill this gap. This paper evaluates performance of three supervised learning algorithms of artificial neural network by creating classifiers for the complex problem of latest web spam pattern classification. These algorithms are Conjugate Gradient algorithm, Resilient Back-propagation learning, and Levenberg-Marquardt algorithm.*

## 1. INTRODUCTION

In this paper we are studying three artificial neural network (ANN) algorithms, which are Conjugate Gradient learning algorithm, Resilient Back-propagation algorithm, and Levenberg-Marquardt algorithm. All of these algorithms are supervised learning algorithms.

We are evaluating performance of these algorithms on the basis of classification results as well as computational requirements in training. The application on which we are evaluating these algorithms is web spam classification.

Web spam is one of the key challenges for search engine industry. Web spam are the web pages that are created or manipulated to lead users from search engine result page to a target web page and also manipulate search engine ranking of the target page.

## 2. RELATED WORK

Svore (2007) [1] devised a method for web spam detection based on content-based features and the rank-time. They used SVM classifier with linear kernel.

Noi (2010) [2] proposed a combination of graph neural network and probability mapping graph self organizing maps organized into a layered architecture. It was a mixture of unsupervised and supervised learning approaches but the training time was computationally very expensive.

Erdelyi (2011) [3] achieved superior classification results in experiment using learning methods LogitBoost and RandomForest with less computation hungry content features. They used 100,000 hosts from WebspamUK2007 and 190,000 hosts from DC2010 datasets and investigated the trade-off between feature generation and spam classification accuracy. They proved that more features improve performance but complex features such as PageRank improves the classification accuracy marginally.

According to Biggio (2011) [4], SVM can be manipulated in adversarial classification tasks such as spam filtering.

Similarly, Xiao (2012) [5] showed that injection of contaminated data in training dataset significantly degrades the accuracy of the SVM.

This is well known that neural network perform better with noisy data. Similar is the case of adversarial dataset of web spam.

## 3. ARTIFICIAL NEURAL NETWORK

An ANN is a collection of simple processing units which communicate with each other using a large number of weighted connections. Each unit receives input from neighbour units or from external source and computes output which propagates to other neighbours. There is also a mechanism to adjust weights of the connections. Normally there are 3 types of units.

- **Input Unit:** which receives signal from external source.
- **Output Unit:** which sends data out of the network.
- **Hidden Unit:** which Receives and sends signals within the network.

Many of the units can work parallel in the system. ANN can be adapted to take a set of inputs and produce a desired set of outputs. This process is known as learning or training. There are two kinds of training in neural network.

- **Supervised:** The network is provided a set of inputs and corresponding output patterns, called training dataset, to train the network.
- **Unsupervised:** The network trains itself by creating clusters of patterns. Here no prior set training data is provided to the system.

### 3.1. Multi-Layer Perceptron (MLP)

MLP is a nonlinear feed forward network model which maps a set of inputs x into a set of outputs y. It has 3 types of layers: input layer, output layer, and hidden layer.

Standard perceptron calculates a discontinuous function:

$$\vec{x} \rightarrow f_{step}(w_0 + (\vec{w}, \vec{x})) \quad (1)$$

smoothing is done using a logistic function to get

$$\vec{x} \rightarrow f_{log}(w_0 + (\vec{w}, \vec{x})) \quad (2)$$
$$\text{where: } f_{log}(z) = \frac{1}{1 + e^{-z}}$$

MLP is a finite acyclic graph where the nodes are neurons with logistic activation. Neurons of $i^{th}$ layer serves as input for neurons of $(i+1)^{th}$ layer. Very complex functions can be calculated with network containing many neurons. All the neurons of one layer are connected to all neurons of next layer.

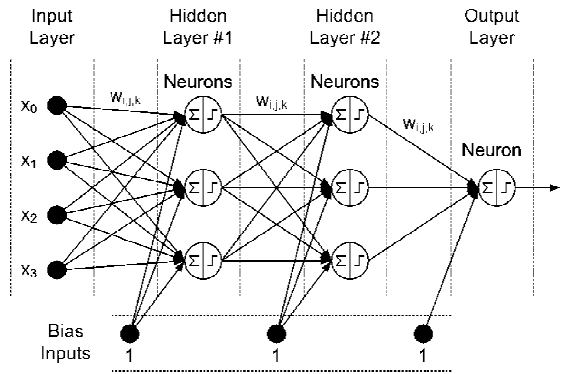

Figure 1. A Multilayer Perceptron with one input, two hidden, and one output layer.

All connections are weighted with real number. Weight of connection $i \rightarrow j$ is $w_{ji}$. All hidden and output layer neurons have a bias weight $w_{i0}$.

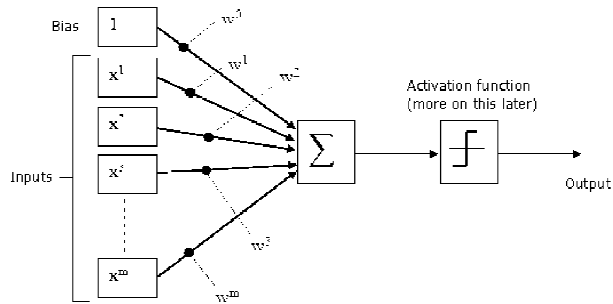

Figure 1. A Neuron outputs an activation function applied over weighted sum of its all inputs

Each node of the network outputs an activation function applied over weighted sum of its all inputs.

$$a_i = f(w_{i0} + \sum_{j=1}^{n} w_{i,j} X a_j) \qquad (3)$$

The network output is given by the $a_i$ of the output neurons.

### 3.2. Neural Network Supervised Learning Algorithms

While implementing any learning algorithm, our objective is to reduce the Global Error E defined by:

$$E = \frac{1}{P} \sum_{p=1}^{P} E_p \qquad (4)$$

Where P is the total size of training dataset, and $E_p$ is the error for training pattern p. And also,

$$E_p = \frac{1}{2} \sum_{i=1}^{N} (O_i - t_i)^2 \qquad (5)$$

where N is total number of output nodes, $O_i$ is output of the i[th] output node and $t_i$ is the target output at the i[th] output node. Every learning algorithm tries to reduce the global error by adjusting weights and biases. Now first we will discuss the three algorithms one by one.

### 3.3. Conjugate Gradient (CG) Algorithm

It is basic back-propagation algorithm. It adjusts weights in the steepest descent direction i.e. the most negative of the gradients. This is the direction in which the function is decreasing most rapidly. It is observed that although the function decreases most rapidly along the negative of the gradient direction, it does not always provide the fastest convergence. In the Conjugate Gradient algorithm a search is done in conjugate directions, which generally provides the faster convergence than the steepest descent direction[6].

Conjugate Gradient Algorithm adjusts step size in each iteration along the conjugate gradient direction to minimize performance function. It searches the steepest descent direction on the first iteration.

$$p_0 = - g_0 \qquad (6)$$

Then it performs line search to determine the optimal distance to move along the current search direction by combining new steepest descent direction with the previous direction.

$$p_k = - g_k + \beta_k \; p_{k-1} \qquad (7)$$

where the constant $\beta_k$ is

$$\beta_k = \frac{g_k^T g_k}{g_{k-1}^T g_{k-1}} \qquad (8)$$

It is the ratio of the norm squared of current gradient to the norm squared of the previous gradient[7].

### 3.4. Resilient Back-propagation (RB) Algorithm

Multilayer Perceptron networks typically use sigmoid transfer function in the hidden layer. It is also known as squashing function because it compresses an infinite input range into finite output range. The sigmoid function's slope approaches to zero as input gets large. It causes problem when we use steepest descent to train network. Because since gradient may have very small value so it may cause small changes in weights and biases even though the weight and biases are fairly far away from the optimal values.

The purpose of the resilient back-propagation (RB) learning is to remove the harmful effect of the magnitude of partial derivatives. It uses only the sign of the partial derivative to determine the direction of the weight update. The magnitude of the weight change is calculated as following rule [8]:

- Values of each weight and bias are increased when the derivative of performance function with respect to that weight has same sign for two successful iterations.
- Values of each weight and bias are decreased when the derivative with respect to weight changes sign from the previous iteration.
- If the derivative is zero the values of the weights and biases are remain same.
- When weights oscillate, values of weight change is reduced.
- If weights continuously change in the same direction, weight change is increased.

### 3.5. Levenberg-Marquardt (LM) Algorithm

It provides a numerical solution to the problem of minimizing a generally non-linear function, over a space of parameters for the function. It is popular alternative to the Gauss-Newton method of finding the minimum of a function. It approaches second order training speed without computing the Hessian Matrix [9],

If the performance function is of the form of a sum of squares, then Hessian matrix can be approximated as:

$$H = J^T J \tag{9}$$

and the gradient is:

$$g = J^T e \tag{10}$$

where J represents the Jacobian matrix containing the first derivatives of network error with respect to weights and biases, and e is the vector of network errors.

The Jacobian matrix can be computed through a standard back-propagation technique that is much complex than computing Hessian matrix. This approximated matrix is used in following Newton like update

$$x_{k+1} = x_k - [J^T J + \mu I]^{-1} J^T e \tag{11}$$

When the scalar $\mu = 0$, it becomes the Newton's method on approximated Hessian matrix. When $\mu$ is large, it becomes gradient descent with small step size. The Newton's method is faster and more accurate so algorithm shifts towards it. So the $\mu$ is decreased with each successful step (i.e. when performance function reduces). It is increased when a tentative step would increase the performance function. So the performance function always reduces at each iteration. It is more powerful than conventional gradient descent technique.

LM Algorithm is very sensitive to initial network weights. It does not consider outliers in the data which may lead to over-fitting noise. To avoid these situations, Regularization technique is used. One of such technique is Bayesian Regularization.

### 3.6. Bayesian Regularization (BR)

It is used to overcome the problem of interpolating noisy data. Mackay (1992) proposed Bayesian framework which can be directly applied to the NN learning problem. It allows to estimate effective number of parameters actually used by the model i.e. the number of network weights actually needed to solve a particular problem. Bayesian Regularization expands the cost function to search not only for minimal error but for minimal error using minimal weights. By using Bayesian Regularization one can avoid costly cross validation. It is particularly useful for the problems that cannot, or would suffer, if a portion of available data were reserved to a validation set. Furthermore, the regularization also reduces or eliminates the need for testing different numbers of hidden neurons for a problem. Bayesian regularized ANN are difficult to be over trained and over-fit.

## 4. CONFUSION MATRIX

Confusion matrix or contingency table is used to evaluate the performance of a machine learning classifier. We used four attributes of confusion matrix while evaluating the performance of the algorithms. These attributes are Sensitivity, Specificity, Efficiency and Accuracy. These attributes are defined as following:

**Sensitivity** or **True Positive Rate (TPR)**, also known as **Recall Rate** is given by:

$$\text{Sensitivity} = \frac{TP}{(TP + FN)} \quad (12)$$

**Specificity** or **True Negative Rate (TNR)** is given by:

$$\text{Specificity} = \frac{TN}{(FP + TN)} \quad (13)$$

**Efficiency** is given by:
$$\text{Efficiency} = \frac{\text{Sensitivity} + \text{Specificity}}{2} = \frac{TPR + TNR}{2} \quad (14)$$

**Accuracy** is given by:
$$\text{Accuracy} = \frac{TP + TN}{(P + N)} \quad (15)$$

where P is number of positive instances, N is number of negative instances, TP is number of correctly classified positive instances, TN is number of correctly classified negative instances, FP is number of incorrectly classified as positive instances and FN is number of incorrectly classified as negative instances.

## 5. EXPERIMENT

We evaluated 3 supervised learning algorithm which uses back-propagation on multilayer perceptron neural network. We checked the performance of these algorithm to automatically detect web spam. These algorithms are Conjugate Gradient, Resilient Back-propagation and Levenberg-Marquardt learning.

We created neural network with single hidden layer and used one neuron in the output layer. We employed bipolar sigmoid function which has the output range of [ -1 , 1 ].

$$f(x) = \frac{2}{1 + e^{-\alpha x}} - 1 \quad (16)$$

We selected in the experiment:

Stopping Criteria as Number of Iterations θ = 100,
Learning Rate α = 0.1

Advanced Computational Intelligence: An International Journal (ACII), Vol.2, No.1, January 2015Number of neurons in hidden layers = 10 (or 20 where specified).

We created a corpus of 368 instances of manually selected web pages, in which about 30% instances were labelled as spam and rest of the pages were labelled as ham. To create training dataset, we randomly selected about 80% of records from the corpus and for testing we used remaining 20% of the records.

We extracted total 31 low cost quality features of the pages and categorized them in 3 categories: URL (10 features), Content (16 features) and Link (5 features). We call these factors low cost because they are computationally less expensive to be extracted. These features are listed in table 1.

Table I. Low Cost Quality Features of a Web Page

| URL Features |
| --- |
| SSL Certificate |
| Length of URL |
| URL is not a sub-domain |
| TLD is authoritative (*.gov, *.edu, *.ac) |
| Domain contains more than 2 same consecutive alphabet) |
| Sub-domain is more than level 3 (e.g. mocrosoft.com.mydomain.net) |
| Domain contains many digits and special symbols |
| IP address instead of Domain Name |
| Alexa Top 500 site. |
| Domain Length |
| **Content Features** |
| HTML Length |
| Word Count |
| Text Character Length |
| Number of Images |
| Description Length |
| Existence of H2 |
| Existence of H1 |
| Video Integration |
| Number of Ads |
| Title Character Length |
| Compression Ratio of Text |
| Ratio of text to HTML |
| Presence of alt text for images |
| Presence of obfuscated JavaScript code (pop-ups, redirections etc) |
| % of Call to Action in the Text |
| Percentage of Stop words in Text |
| **Links Features** |
| Number of Internal links |
| Internal link is self referential |
| Number of External links |
| Fraction of anchor text / Total Text |
| Word Count in Anchor Text |

27

To test the performance of each algorithm, we trained, tested and obtained 10 performance result values for each category and calculated the average.

We used Accord.net and Aforge.net libraries to create neural network.

## 6. RESULT

We have created tables (Table II to Table VIII) to show the performance results of each algorithm. Consider the number of neurons in hidden layer as 10 unless specified. The values in the tables represent average of 10 experimental readings of each category. The values in underlined show the best results.

Table 2. Using Url Features Only (10 Features)

| Algorithm | Sensitivity (TPR) | Specificity (TNR) | Efficiency | Accuracy | Training Time (seconds) |
|---|---|---|---|---|---|
| CG | 0.9153 | 0.3088 | 0.6121 | 0.5716 | 0.149 |
| RB | 0.4653 | 0.9117 | 0.6885 | 0.7183 | 0.168 |
| LM | 0.6884 | 0.7647 | 0.7265 | 0.7316 | 2.723 |
| LM+BR | 0.4224 | 0.9264 | 0.6744 | 0.7090 | 5.790 |

Table 3. Using Content features Only (16 Features)

| Algorithm | Sensitivity (TPR) | Specificity (TNR) | Efficiency | Accuracy | Training Time (seconds) |
|---|---|---|---|---|---|
| CG | 0.7384 | 0.8823 | 0.8104 | 0.8200 | 0.198 |
| RB | 0.7269 | 0.9205 | 0.8237 | 0.8366 | 0.191 |
| LM | 0.6615 | 0.8441 | 0.7528 | 0.7650 | 9.685 |
| LM+BR | 0.6116 | 0.9380 | 0.7748 | 0.7952 | 16.445 |

Table 4. Using Links features Only (5 Features)

| Algorithm | Sensitivity (TPR) | Specificity (TNR) | Efficiency | Accuracy | Training Time (seconds) |
|---|---|---|---|---|---|
| CG | 0.9230 | 0.6382 | 0.7806 | 0.7616 | 0.113 |
| RB | 0.6692 | 0.9117 | 0.7904 | 0.8066 | 0.157 |
| LM | 0.7230 | 0.9088 | 0.8159 | 0.8282 | 1.923 |
| LM+BR | 0.7038 | 0.8880 | 0.7959 | 0.8080 | 2.722 |

Table 5. Using URL + Links Features (15 Features)

| Algorithm | Sensitivity (TPR) | Specificity (TNR) | Efficiency | Accuracy | Training Time (seconds) |
|---|---|---|---|---|---|
| CG | **0.**9538 | 0.7970 | 0.8754 | 0.8650 | 0.191 |
| RB | 0.8076 | 0.9823 | 0.8950 | 0.9066 | 0.187 |
| LM | 0.8115 | 0.9352 | 0.8734 | 0.8816 | 3.744 |
| LM+BR | 0.9115 | 0.9558 | 0.9337 | 0.9366 | 13.458 |

Table 6. Using URL + Content features (26 Features)

| Algorithm | Sensitivity (TPR) | Specificity (TNR) | Efficiency | Accuracy | Training Time (seconds) |
|---|---|---|---|---|---|
| **No. of neurons in hidden layer=10** | | | | | |
| CG | 0.8846 | 0.9264 | 0.9055 | 0.9083 | 0.266 |
| RB | 0.8538 | 0.9764 | 0.9151 | 0.9233 | 0.190 |
| LM | 0.8807 | 0.8823 | 0.8815 | 0.8816 | 10.743 |
| LM+BR | 0.7115 | 0.9911 | 0.8513 | 0.8700 | 59.957 |
| **No. of neurons in hidden layer=20** | | | | | |
| CG | 0.8730 | 0.9558 | 0.9144 | 0.9200 | 0.539 |
| RB | 0.8576 | 0.9823 | 0.9200 | 0.9283 | 0.312 |
| LM | 0.8884 | 0.9205 | 0.9045 | 0.9066 | 32.789 |
| LM+BR | 0.7692 | 1.000 | 0.8846 | 0.9000 | 304.134 |

Table 7. Using Content + Links features (21 Features)

| Algorithm | Sensitivity (TPR) | Specificity (TNR) | Efficiency | Accuracy | Training Time (seconds) |
|---|---|---|---|---|---|
| CG | 0.7269 | 0.8705 | 0.7987 | 0.8083 | 0.230 |
| RB | 0.7593 | 0.9088 | 0.8340 | 0.8438 | 0.188 |
| LM | 0.7038 | 0.8441 | 0.7739 | 0.7833 | 7.940 |
| LM+BR | 0.6846 | 0.9470 | 0.8158 | 0.8333 | 36.674 |

Table 8. Using URL + Content + Links features (31 Features)

| Algorithm | Sensitivity (TPR) | Specificity (TNR) | Efficiency | Accuracy | Training Time (seconds) |
|---|---|---|---|---|---|
| **No. of neurons in hidden layer=10** | | | | | |
| CG | 0.8500 | 0.9676 | 0.9088 | 0.9166 | 0.297 |
| RB | 0.8769 | 0.9735 | 0.9252 | 0.9316 | 0.214 |
| LM | 0.8653 | 0.9029 | 0.8841 | 0.8866 | 13.146 |
| LM+BR | 0.6923 | 0.9852 | 0.8388 | 0.8583 | 92.248 |
| **No. of neurons in hidden layer=20** | | | | | |
| CG | 0.8384 | 0.9823 | 0.9104 | 0.9200 | 0.575 |
| RB | 0.8807 | 0.9588 | 0.9197 | 0.9250 | 0.377 |
| LM | 0.8653 | 0.9264 | 0.8959 | 0.9000 | 45.499 |
| LM+BR | 0.8115 | 0.9882 | 0.8998 | 0.9116 | 357.580 |

Data from the tables suggest that Conjugate Gradient (CG) Algorithm gave best TPR (Sensitivity) in most of the categories, whereas Levenberg-Marquardt algorithm with Bayesian Regularization (LM+BR) gave best TNR (Specificity) in most of the categories.

The overall best classification performance was achieved in most of the categories by Resilient Back-propagation (RB) algorithm in both Efficiency and Accuracy measures.
The training time was found lowest in most of the cases for Resilient Back-propagation (RB) so we can say that it is not only best in classification, it is fastest algorithm as well. Conjugate

Advanced Computational Intelligence: An International Journal (ACII), Vol.2, No.1, January 2015

Gradient (CG) works fast when number of inputs are less, but when number of inputs are increased, its becomes slower than RB. The slowest algorithm observed was Levenberg-Marquardt with Bayesian Regularization (LM+BR) in all of the cases in the experiment.

The overall classification performance of each algorithm improved with increased number of page quality factors but training time also increased. The cases where the number of quality factors were high, increasing number of neurons improved the classification performance of algorithms but it also increased the training time.

## 7. CONCLUSION

In the experiment we can conclude that the Resilient Back-propagation algorithm is fastest and performs best in both Efficiency and Accuracy measures. Conjugate Gradient algorithm gives best sensitivity, whereas Levenberg-Marquardt algorithm with Bayesian Regularization gives best specificity but it is the slowest when training time is considered.

Classification performance of each algorithm improves with increased input factors. If the number of factors are high, increased number of neurons improves the performance of algorithm. The training time increases for all algorithms when either number of inputs are increased or number of neurons in hidden layer are increased.